# More efficient manual review of automatically transcribed tabular data

Bjørn-Richard Pedersen[1*], Rigmor Katrine Johansen[2], Einar Holsbø[3], Hilde Sommerseth[1], Lars Ailo Bongo[3]

[1] Norwegian Historical Data Centre, UiT The Arctic University of Norway
[2] Department of Health and Care Sciences, UiT The Arctic University of Norway
[3] Department of Computer Science, UiT The Arctic University of Norway

* Corresponding author: bjorn-richard.pedersen@uit.no

## Abstract

Machine learning methods have proven useful in transcribing historical data. However, results from even highly accurate methods require manual verification and correction. Such manual review can be time-consuming and expensive, therefore the objective of this paper was to make it more efficient.

Previously, we used machine learning to transcribe 2.3 million handwritten occupation codes from the Norwegian 1950 census with high accuracy (97%). We manually reviewed the 90,000 (3%) codes with the lowest model confidence. We allocated those 90,000 codes to human reviewers, who used our annotation tool to review the codes. To assess reviewer agreement, some codes were assigned to multiple reviewers. We then analyzed the review results to understand the relationship between accuracy improvements and effort. Additionally, we interviewed the reviewers to improve the workflow.

The reviewers corrected 62.8% of the labels and agreed with the model label in 31.9% of cases. About 0.2% of the images could not be assigned a label, while for 5.1% the reviewers were uncertain, or they assigned an invalid label. 9,000 images were independently reviewed by multiple reviewers, resulting in an agreement of 86.43% and disagreement of 8.96%.

We learned that our automatic transcription is biased towards the most frequent codes, with a higher degree of misclassification for the lowest frequency codes. Our interview findings show that the reviewers did internal quality control and found our custom tool well-suited. So, only one reviewer is needed, but they should report uncertainty.

# Introduction

The Norwegian Historical Data Centre at UiT the Arctic University of Norway is one of the partners in constructing the Norwegian Historical Population Register (HPR) (http://www.rhd.uit.no/nhdc/hpr.html). HPR will include the records of 9,7 million people who lived in Norway in the period from 1801 to 1964, and the main building blocks are population censuses and church books. One important goal for HPR is to transcribe historical population sources with a high degree of accuracy, and machine learning can be a key tool in achieving this while keeping costs and time usage low. The building blocks in HPR have been transcribed using three approaches. First, professional transcribers and volunteers have manually transcribed about 30 million person entities from 19th and early 20th century population censuses and church books that date back to the early 18th century. Second, selected columns in the church books have been manually transcribed through an agreement between the National Archives of Norway and three commercial genealogy companies. Third, we have developed a machine learning model to automate the transcription of 2.3 million handwritten occupation codes from the Norwegian 1950 population census (Pedersen et al. 2022). The HistLab at the NHDC will, in close collaboration with the Health Data Lab at UiT https://uit-hdl.github.io/, continue to use machine learning for transcription of additional census columns and other historical data.

We had three main requirements for the automated occupation code transcription: (i) the accuracy of the automatic transcription should be similar to human accuracy (97%), (ii) given the 7.3M images we extracted from the 1950 census, fewer than 100,000 images should be sent for manual validation and correction, and (iii) there should be no systematic errors introduced to the automatically transcribed data. There is a trade-off between the first and second goals, since a higher accuracy requires more manual transcription. We can tune this trade-off by adjusting the prediction confidence required for a code to be automatically transcribed. A higher confidence threshold improves accuracy, but requires more manual validation and correction. But this also means that more efficient manual work can increase machine learning accuracy. An important part of tuning a machine learning model for transcription is therefore to find the best trade-off between these two conflicting goals.

Our resulting model satisfied all three requirements. We achieved 97% accuracy, but for 82,177 images the classification confidence score was below the 65% threshold we had set as a criteria (Pedersen et al. 2022). In addition, during post-processing of the results, we found 16,156 images that had been given invalid or nonsensical codes. Since we have to transcribe all images, these 98,333 images must be manually validated and possibly corrected. This manual review requires a significant amount of human effort and time. For other data, with non-numeric columns that are more challenging for machine learning, we expect lower model accuracy and therefore even more manual work to maintain the required accuracy for the automatically transcribed images. It is therefore important for transcription projects to make this time-consuming and costly manual work efficient.

The need for human labeling in machine learning projects have motivated the development of data labeling tools such as Amazon Mechanical Turk (https://www.mturk.com/), Label Studio (https://labelstud.io/), Prodigy (https://prodi.gy/), Snorkel Flow (https://snorkel.ai/snorkel-flow-platform/), and many more (https://github.com/doccano/awesome-annotation-tools provides a list). However, the tools may have an expensive license since they are often targeted for the needs of large companies or organizations. Typically these tools aim to reduce the effort of labeling the data required to train machine learning models. They are therefore designed for efficient labeling of training data by a team of laborers. Many enable the use of humans-in-the-loop when training a machine learning model by for example using active learning (Bernard, Zeppelzauer, et al. 2018) or programmatic labeling (Cohen-Wang et al. 2019). Also, with the recent focus on data-centric machine learning (https://datacentricai.org/), an important aim is to improve the quality of the labels. For example, by analyzing and flagging labels that differ between human annotators. Most commercial tools therefore support such quality control workflows. In addition, there are also specialized tools for data cleaning (such as Trifacta, https://www.trifacta.com/), and for correcting labels (Xiang et al. 2019). Finally, the field of explainable AI (Dwivedi et al. 2023; Kim et al. 2023) is focusing on understanding the results of machine learning models and have therefore developed many libraries and tools especially for image analysis (https://github.com/wangyongjie-ntu/Awesome-explainable-AI provides lists of tools). It is therefore a need to understand how these tools can be used in transcription projects and especially for the necessary manual verification and correction. The manual correction dataset is typically much larger than the training dataset and it is more noisy. Consequently, it is more challenging and time-consuming to label these with high accuracy.

High quality manual review of labels assigned by machine learning models can be a tedious and boring task. It is therefore also important to understand how the workflow for the human reviewers can be structured. In addition, professional transcribers differ from domain experts that do data labeling ad-hoc and gig-workers that label many different types of data, in that they are trained to interpret historical documents and ensure that the transcription is accurate. They may therefore have different needs for tool support and workflow organization.

To address the challenge of making manual review efficient, accurate, and motivating for historical data reviewers, we used a custom annotation tool to verify and correct the Norwegian 1950 census occupation codes that our model could not automatically transcribe with confidence to get insight into how accurate and time-consuming this manual work is. Then, we interviewed the reviewers to understand how to improve the manual workflow.

# Methods

## Pilot study

We conducted a pilot study to identify i) which features of state-of-the-art data labeling tools are useful for this task, ii) which potential problems to be aware of when performing the manual review, and iii) how to analyze the results after the end of the project period.

## Dataset for manual verification and correction

We developed a machine learning method for automatic transcription of occupation codes in the Norwegian 1950 census (Pedersen et al. 2022). Here we used the labeled codes resulting from using the method on the entire Norwegian census of 1950's occupational code column, which consists of 7,342,113 images. 82,177 images did not pass the confidence threshold we had set for automatic transcription, 2,273 images had a nonsensical label (e.g. 't4b'), 10,376 images had labels that were not in Statistics Norway's official list of occupation codes for the 1950 census, 3,507 images had labels that exist in the official list, but were not a part of our training set, and as such we decided to include them in the manual review. Of these 98,333 images, we used 8,333 for validation tests and for the pilot study, so there were 90,000 codes to review in this study.

## Reviewers

We recruited six employees from the Norwegian Historical Data Centre to perform the manual review of our images. They are all women and three of them are professional transcribers, holding several years worth of experience with transcription of historical population sources. The other three were research assistants/students who have been working with historical data, but never with this type of handwritten source material. In addition, one of the authors, Bjørn-Richard Pedersen (B.R.P.), contributed as a reviewer. We believe these seven reviewers represent levels of experience that are realistic for this type of project.

We divided the above 90,000 unclassified images evenly among the seven reviewers, yielding roughly 13,000 images each. They were given one month to review these images.

## Custom annotation tool used to review images

In our previous paper (Pedersen et al. 2022) we manually labeled a training set using an annotation tool that we created specifically for manual verification and correction of labeled images (https://github.com/HistLab/More-efficient-manual-review-of-automatically-transcribed-tabular-data). We used the same tool in this project. The graphical user interface of the tool shows up to 60 images at once, grouped by the prediction made by the machine learning model. The model prediction is shown at the top of the window (Figure 1). The reviewers manually reclassify the labels that were wrong by entering the new label into the textbox for the corrected image. The correctly labeled images are ignored in this process.

The manual reclassification should adhere to the Histform-standard (https://www.digitalarkivet.no/en/content/guidelines-histform). This standard consists of a set of guidelines that transcribers should follow when transcribing historical sources, but are also suited for the reviewing process in this project, among these are a set of specific symbols to use when the reviewer is uncertain ('??'), and when the original text has been crossed out and replaced with new text ('<new text> %<old text>%'), among others. Because our reviewers also included students we created a set of instructions that summarize the relevant parts of the Histform-standard. These instructions were handed out to each reviewer before they started.

To understand the manual effort required for this project, we modified the tool to record the time used by each reviewer for each group of images. We removed the time records where it seemed obvious that the reviewer had taken a break, as there was no pause function in the tool. The reviewers were aware that they were being timed, and we also discussed this in the interviews. We anonymized all data and stored the images, corrected labels, and per-group times in a single database.

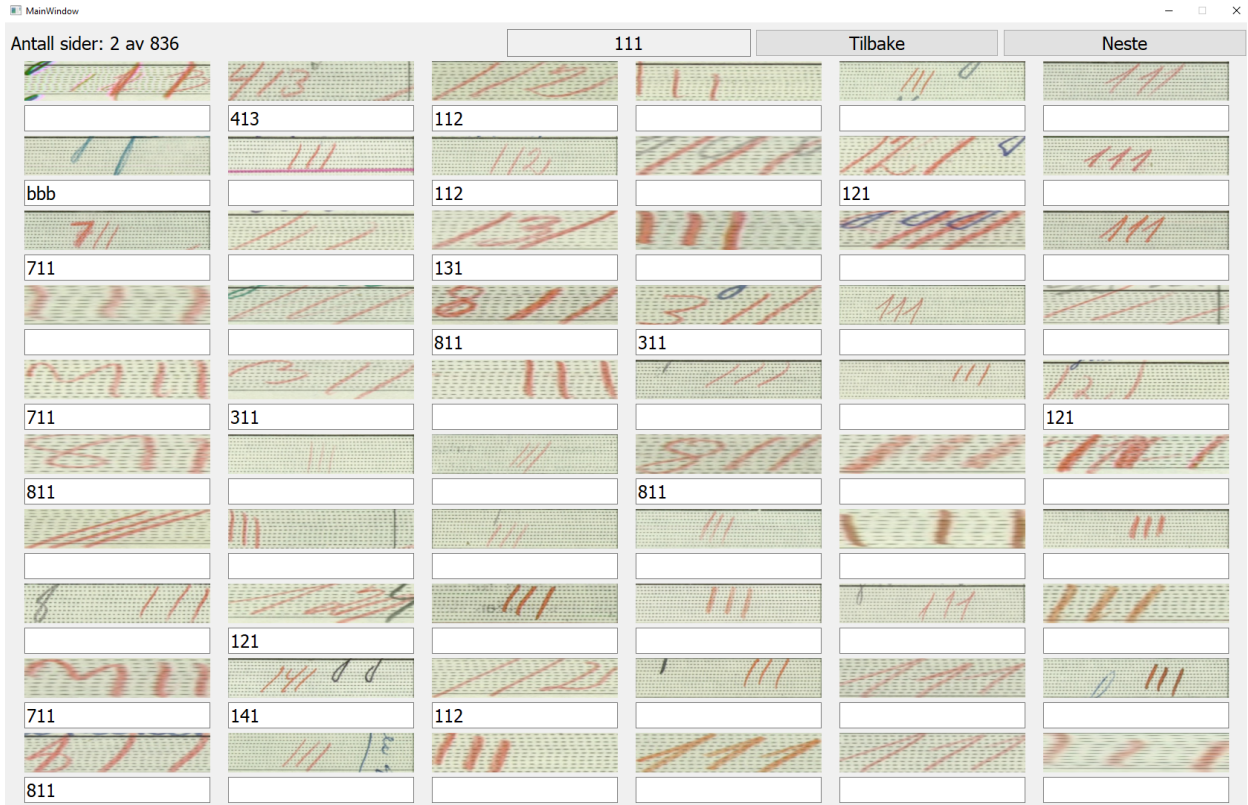

*Figure 1:* A screenshot of our custom annotation tool used for the manual review. The text on the upper left translates to "number of pages: 2 of 836". The buttons on the right translate to "back" and "next".

## Misclassifications done by the machine learning model

The distribution of occupation codes may affect the trained model. Specifically, the 11 most common occupations comprise 70% of the images in our data, while the remaining 30% of images belong to one of 329 other occupations. Some codes are seen only once. Hence the distribution of codes in these data is highly skewed, and the model gets to train on thousands of examples of certain codes and may see other codes only once or not at all. There is a question whether these small occupations get misclassified more often and, if so, whether they get misclassified as more common occupations.

Performing a manual review after the classification allows us to discover how big the misclassification-rate of our model per class is, and to also understand which classes many codes are incorrectly labeled as. For the analysis, we removed human labels that contain uncertainty symbols. We therefore assume that the remaining human labeled codes are correct. We then separated the remaining labels into two categories: where human and machine agreed, and where they disagreed. We can then count the number of instances for each label and order our classes by size. To find the classes the model is biased towards, we order the incorrect labels by frequency.

## Verification and correction quality

We randomly selected 10% of the 90,000 automatically labeled code images and assigned them to pairs of two reviewers. We used the results from these overlapping reviews to evaluate the inter-reviewer agreement. In the analysis we ignore the model's labels since we assume that the human reviewers are correct. We therefore use 3 categories (illustrated in table 1):

1. *Certain.* Both of the manual reviewers agreed with each other.
2. *Unknown.* The manual reviewers disagreed with each other.
3. *Uncertain.* The manual reviewers noted down an uncertainty symbol, as defined in the Histform-standard, to indicate that they were not entirely confident in the label they assigned to the image.

We also measured the consistency of the reviewer's own classifications by assigning about ~14% of each reviewer's images back to themselves. These images were not used when calculating inter-reviewer performance, but rather as a test of the individual reviewer's attention to their task. We wanted to observe if they consistently labeled the images.

*Table 1:  An overview of the categories we use in the project, and examples for how an occupation code image may be categorized based on the labels from the human reviewers.*

| Reviewer 1 | Reviewer 2 | ML model label | Category |
|---|---|---|---|
| 531 | 531 | 531 | Certain |
| 531 | 531 | 999 | Certain |
| 531 | 999 | 888 | Unknown |
| 531 | 888 | 888 | Unknown |
| 531 | 182?? | 141 | Uncertain |
| 531@533 | 531 | 181 | Uncertain |
| ?? | 531 | 555 | Uncertain |
| 1??8 | 531 | 188 | Uncertain |

## Interviews and analysis

We used qualitative interviews to gain insight into how each reviewer experienced the work, workflow, and the use of our custom tool. When the seven reviewers (including

B.R.P.) had completed their manual verification and correction work, Lars Ailo Bongo (L.A.B.) and B.R.P. interviewed the other six reviewers, utilizing a semi-structured interview guide prepared with four main questions to ensure the acquisition of data believed most interesting for this project. The interviews were performed in Norwegian and the interview guide questions translated to English were:

1. How did you experience the task?
2. Did you find any particular part more challenging than the rest?
3. What are your impressions of using the custom annotation tool?
4. You were informed in advance that you would be timed when performing this task, did that impact your work in any way?

We also asked follow-up questions when relevant. During the interviews we took notes, and after each interview the notes were reread and additional information was added. We also discussed the interviews and shared our first impressions of them, in particular when it came to the topics and themes that seemed most important to the individual reviewers. Finally, once all interviews were conducted we analyzed the compiled notes, drawing out what seemed to be the key findings for our aim.

### Research ethics

Before the research project commenced B.R.P and Hilde Sommerseth (H.S.) arranged a meeting on Zoom with the six reviewers who had orally consented to participate in the project. We informed them about the project goals, that their task would be to transcribe approximately 13,000 handwritten occupation code images from the 1950 census into digital text, and of practical details about the project. The resulting data each reviewer produced was anonymized and aggregated. We also informed them that L.A.B and B.R.P would interview them once the manual work was completed. All of them agreed to this and were given the opportunity to ask clarifying questions. The participants knew who participated, hence there was no anonymity among the reviewers. To ensure anonymity for the six reviewers who were interviewed, the interview data was only available to the interviewers and those authors of the paper necessary for analysis of the interviews.

## Results

### Corrected machine learning labels

Of the 90,000 images in the dataset, our reviewers corrected the model labels in 62.8% of cases. They agreed with the model in 31.9% of cases. The reviewers were not able to confidently label 0.2% of the images, and 4.8% of the images were given some other type of uncertainty symbol. We believe most of these can be programmatically resolved. For example, if reviewer A labeled the image '531' and reviewer B labeled it '531@537', we would be fairly confident that the label should be '531'. For the remaining 0.3% of the images, the reviewers had given an invalid input, such as writing two codes in the same text box, or agreed with the model on a label that could not possibly be true, such as '5bb'.

## Verification and correction quality

We found that 86.4% of the images reviewed by two reviewers were categorized as *Certain*. Both reviewers agreed with the machine learning label for 33.83% of the images, which is similar to the results above. 8.9% are categorized as *Unknown*; for 44.89% of these, one reviewer agreed with the machine learning label. 4.7% belonged to the *Uncertain* category.

The results show that in 8.9% of cases the reviewers disagreed with each other, and at least one of them was therefore wrong. Since the images in this manual verification dataset are some of the hardest images to label, we conclude that an error rate of 8.9% is good enough, and so a second reviewer is not needed.

We also found that in these overlapping reviews, only 34.04% of images had at least one reviewer agreeing with the model's labels. We believe these results combined show that these particular images are challenging to label, both for humans and the machine learning model.

## Labeling consistency

Each reviewer got, on average, 179 images that were duplicates taken from their own sample of images. Three of the reviewers gave exactly the same label to all of their images, and the other four reviewers had two to three images where they labeled differently. Some of these differences were made up of labels where the reviewer didn't enter the second label fully, e.g. the first label was written down as '555' and the second label was written as '55', or where the reviewer reordered the label, writing '861' for the first label and '168' for the second. The results therefore show that the reviewers are consistent.

## Model classification error analysis

The misclassification rate of our model is larger for the smaller classes (Figure 2). For some of the smallest classes, the model has a misclassification rate of 100%. The same is true for two of the largest classes as well ('blank' and 'text'), these were included in the model to handle cases where the process of extracting images from the census pages did not manage to cut only images containing a code. In this dataset, the model always predicted some combination of 'blank'/'text' and digits for these images, most likely because of the noisy nature of an incomplete image extraction, and never just 'blank' or 'text'.

We analyzed the labels the model assigned incorrectly to the images, as well as the most frequent labels in the training set. The reviewers have written 549 unique codes. The training set has 286 codes, and there are 339 official occupation codes. There is some overlap (Figure 3). Four of the largest classes in the training set '531', '899', '555', and '111' are all among the most frequently misclassified by the model, as expected since we assume the model is more likely to predict a large class. But, the smallest classes show a different pattern. After examining two of the smallest classes, '324' and '333' where the model had a 100% misclassification rate, we found that the model would most often predict a label that resembled the codes in appearance or were made up of the same individual digits but in a different order. To illustrate, for the code '324' the model predicted, from most often to less often: '224', '334', '384', or '534', among others. For the code '333', the model would predict: '533', '535', '353', or '323', among others.

We investigated which codes the model incorrectly classified as one of the four largest classes in the training set (Figure 4). Digit similarity is one of the main indicators for the model's labels, but we also find that several of the images that were misclassified as some of the biggest classes were in fact images that should have been labeled 'blank' or 'text' ('bbb' and 'ttt' respectively in the figures). This could indicate that, if the model was unsure about a label it would try to find the closest match, but if there were no actual digits in the image to base this comparison on the model defaulted to a big class.

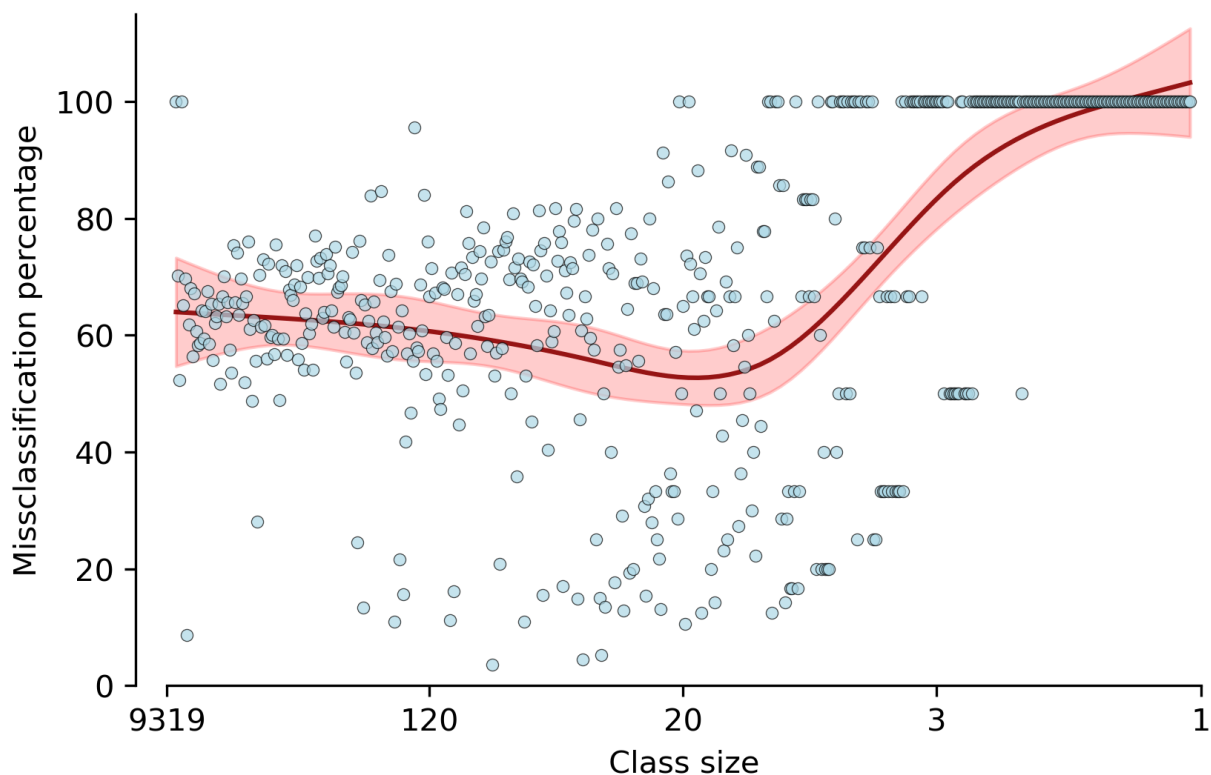

*Figure 2*: Scatter plot for the classification error rate of the model. The x-axis represents all classes ordered by size, with some class size values shown. The regression is a natural spline with six knots that shows the general trend. Note that we assume that all human labels are correct.

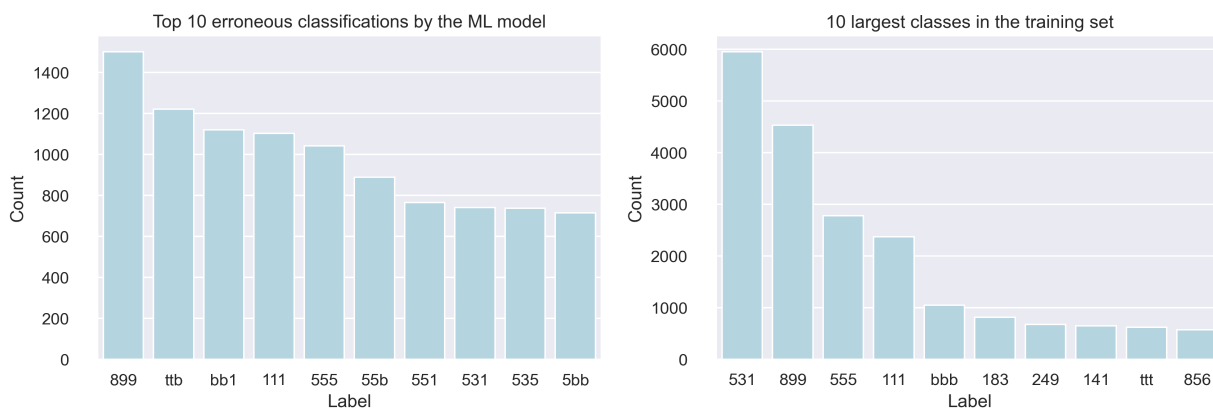

*Figure 3:* *The 10 most frequent misclassified label classes (left), and the 10 most frequent classes in the training set (right).*

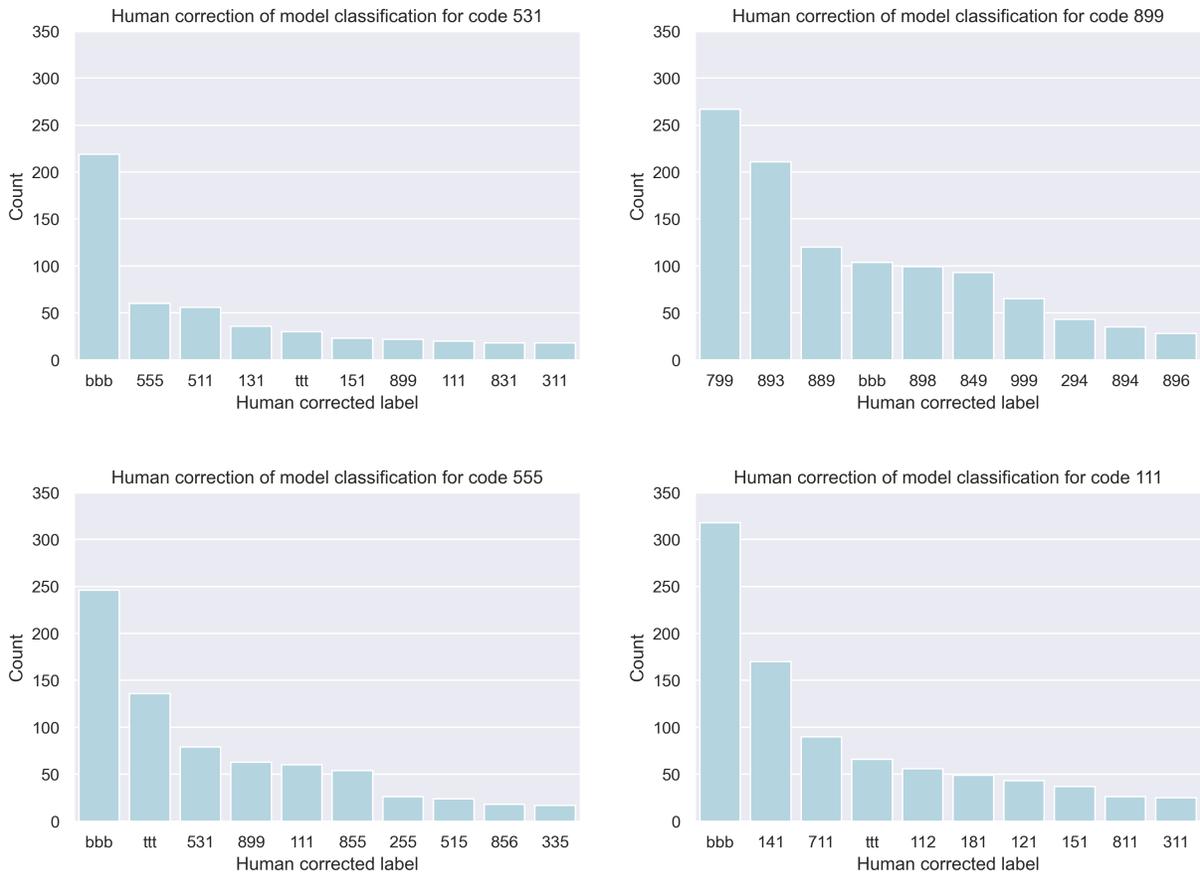

*Figure 4:* *Re-classifications done by the reviewers for the four biggest classes in the training set for the model.*

## Time usage

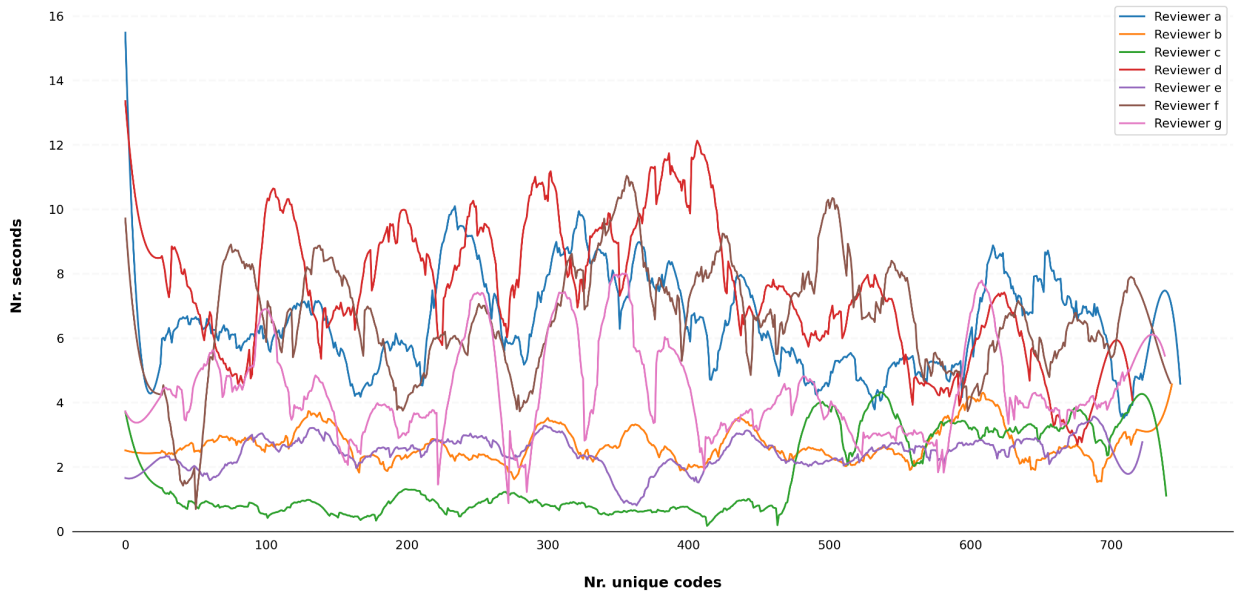

*Figure 5: Time required per reviewer for the start (left side of figure), middle, and end (right side of figure) of the project period. Note that, while all reviewers got the same amount of images, they did not get the same number of codes in each class.*

We examined whether the time spent on reviewing the images changed over the course of the task, but we did not find a clear trend (Figure 5). As expected, some reviewers spent more time at the start of the project period, most likely to get accustomed to the annotation tool. The results suggest that the reviewers do not spend less time at the end of the tasks, and therefore indicate that the quality of the labels is consistent over time.

## Interview findings

The reviewers experienced the task differently. Several found the task tedious and not something they would like to do often. However, they did not mind doing it occasionally or in shorter intervals as a break from their regular tasks of transcribing handwritten text in church books or other census information. Some felt that they contributed to an important and relevant project which in the future may impact their work. Some also found the task very enjoyable and fun, and also said it suited their personality better than their other tasks, and said they could happily do this type of work more often. We believe that the reviewers' positive attitudes reflect that we succeeded in informing and involving the reviewers in this research project.

The reviewers reported that the main challenge in the labeling was caused by inaccuracies in the image extraction algorithm, leading some images to not be correctly cut from the census pages, so the missing parts made it harder and more time-consuming to interpret. Other issues consisted of codes that had been rendered unreadable due to excessive correction or overlapping writing, and codes that were too similar to accurately be told apart. The professional transcribers are used to reverting to the uncut source when they felt it necessary, and therefore missed this opportunity in the annotation tool.

We found that some reviewers had an internal quality control when they encountered images that were hard to discern. The professional transcribers discussed these challenging images with their colleagues, as they do when transcribing other sources. We did not find the same behavior among the less experienced reviewers. However, some thought that they were not allowed to do so since this was a research project. None of the reviewers were concerned about being timed when performing the manual review. Instead they focused on making sure their labels were correct, thereby prioritizing quality over quantity. This is an important finding, since it strongly suggests that if quality control is a part of the work culture there is no need to implement a process for quality control, and therefore also not a need for tool support for such a process.

All reviewers had positive experiences using the custom annotation tool, and felt it was fast and easy to use. The only feature they requested was a clearer indicator of overall progress, such as a line of text in the tool specifying the number of images completed and remaining. We believe that this shows that it is possible to use a special purpose tool that can be implemented in a short amount of time.

Almost all reviewers said it was easy to determine if the code in the image differed from the model's label, at least in cases where the code was not obstructed from view. They were confident in their own relabeling, and would usually do a final overview of all the images on the page before moving on to the next after they were done with the current page. However, one reviewer was concerned that she may be influenced by the suggested labels by the machine learning model.

During the interviews most reviewers talked about images they had found challenging to code, and expressed concerns about their interpretation of the Histform coding rules. They also commented which numbers and codes the automatic translation seemed to make mistakes for. We believe this first point shows a need for better training for the reviewers before starting the task, for example by doing a tutorial. The second point shows that the reviewers can provide input to an error analysis or help tune the model.

When asked if they thought they had worked more efficiently in the beginning, middle, or end of the project period, several of the reviewers felt that they worked more efficiently towards the end of the project. They said that it took a while to understand how the Histform standard's rules would apply to the different images. Once they learned the rules, they felt their pace increasing. However, in the time measurements we do not see this trend (Figure 5). The reviewers believe that they could have worked faster if they were told that speed was important, but most felt that they wanted to put the emphasis on the quality of the work. We therefore believe that the quality of the reviews is consistently high.

# Discussion

## Main findings

Seven reviewers examined a total of 90,000 occupation code images using our custom annotation tool. 9,000 of these were reviewed by two people. 62.8% of the images were corrected. For images with two reviewers, they agreed for approximately 90% of the images. This shows that a better use of human resources could be to have fewer reviewers conducting the manual review with no overlapping images, and any extra resources could be spent on annotating a larger training set during the beginning of a project.

We found that our model is biased towards classes with a higher frequency, and has a misclassification rate up to 100% for the smallest classes. This proves that it is important to conduct this kind of manual validation and correction check, especially for imbalanced datasets in which the smallest classes have few samples. Without a manual review, several occupations would have vanished from the official records.

We explored the users' experiences of the work and the custom annotation tool using semi-structured qualitative interviews. We learned from the interviews that the reviewers prioritize quality over speed, and that some of the reviewers perform an internal quality control as part of their normal tasks. We believe that this behavior should be encouraged among all reviewers, especially if those with less experience are working together with the more experienced ones. This further reduces the need for multiple persons per image. In addition, we found that the reviewers can record uncertainty as part of the labeling task (as also suggested in (Lu, Chang, and Igarashi 2022)). This uncertainty can be used for additional control of the most difficult analysis, or by the researchers analyzing the transcribed data. We also found that even if the task itself is perceived as boring, doing it as part of a short term project is meaningful and interesting. Finally, we found that our custom tool had the functionality and user-friendliness needed for the task ((Zhang et al. 2022) describes a system for building such custom tools). We therefore do not see a need to use commercial labeling tools. Finally, we believe that the interviews show the importance of close collaboration between the machine learning developers and the annotators.

## Related work

There are few research articles on manual review and correction of machine learning classified images and to our knowledge, this is the first study on the evaluation of machine learning classification results for specifically historical population sources. We can therefore not directly compare our results. However, we can compare our approach with the three generic steps for verifying and correcting a training dataset. First, the images that are likely to be mislabeled are identified. We use the model confidence score to detect potentially mislabeled images since we already have trained an accurate model. Without a classification model, clustering methods or active learning can be used. Second, the selected images are manually verified and corrected. We use a custom annotation tool that groups the images according to the predicted label. There are however more advanced algorithms for autocorrecting large sets of images (Hung et al. 2015; Liu et al. 2019; Xiang et al. 2019). Third, the corrected labels can be used to find and correct additional labels. We do not believe our task will benefit from such additional correction, since our error analysis results show that the reviewed images are already selected based on specific features.

## Limitations

Our review results show that there are many errors for the smallest classes. However, we only reviewed the images for which the machine learning model had lowest confidence, or were assigned an invalid label. It may therefore also be interesting to do a similar evaluation of the errors and human effort for additional images. For example, the images with 3-6% lowest confidence and a randomly selected 3% of images with high confidence.

The occupation codes were added to the original census sheets by Statistics Norway after the census had been concluded. They also summarized all occupation codes used in a list. We did not constrain our reviewers to follow this list, since we did not want to introduce bias and because we have found that the list is incomplete.[1]

We have not used the corrected labels to retrain the machine learning model. We have also not evaluated an iterative review process where the reviewers use an active learning approach (Bernard, Hutter, et al. 2018).

---

[1] We have not been able to find the reason for why the extra occupation codes were added to the census.

The findings from the interviews are limited to only reflect the views of the six reviewers that participated in this study. However, the themes we found important in the interviews may be expressed by just one person, but still be relevant for the interview findings.

# Conclusion

In this article we conducted manual validation and correction of 90,000 images that had been automatically transcribed by a machine learning model. The aim was to find a simple and efficient way of performing this manual work, while maintaining high accuracy for the image labels, as they will be added to the Norwegian Historical Population Register. We learned the following lessons which we believe can be used as guidelines for efficient manual verification and correction in other transcription projects that use machine learning:

1. The model's classifications are biased towards the largest classes, with the misclassification rate increasing for smaller classes, so manual verification and correction is necessary to preserve the smallest classes.
2. The reviewers can be trained to use an inherent quality control during the verification process, and they should encode uncertainty into the labels. For this type of image classification one person per image is enough, since we proved that inter-reviewer agreement is high. Therefore, the additional multi-annotator features of commercial state-of-the-art data labeling tools are not needed, and custom annotation tools can be quickly implemented and used instead.
3. The reviewers should be involved in the planning of the work so that the tasks are experienced as an interesting break from their usual routine tasks. But the amount of work should not be overwhelming, and it should be easy to see progress in the form of percentage of images that are done within the graphical user interface.

# Declarations

## Acknowledgments

Thanks to the Målselv transcription team at the Norwegian Historical Data Centre for their enthusiastic participation in the project.

## Data availability

The 90,000 reviewed images and labels are open access (CC0 license). They are available at : https://doi.org/10.18710/LYXKN1

The manually annotated occupation code training dataset is open access (CC0 license) and available at: https://doi.org/10.18710/7JWAZX

## Code availability

The review tool code is open sourced using the MIT license. It is available at: https://github.com/HistLab/More-efficient-manual-review-of-automatically-transcribed-tabular-data

## Sources of funding

This work was funded by UiT the Arctic University of Norway through the interdisciplinary strategic project High North Population Studies and funding provided by the Norwegian Research Council (project number 322231).

19